%% file: acl2022.tex
\pdfoutput=1

\documentclass[11pt]{article}

\usepackage[final]{acl}
\usepackage{times}
\usepackage{latexsym}
\usepackage{algorithm}
\usepackage{algpseudocode}
\usepackage[T1]{fontenc}
\usepackage{fixltx2e}
\usepackage[utf8]{inputenc}
\usepackage{microtype}
\usepackage{inconsolata}
\usepackage{graphicx,subcaption} 
\usepackage{array}
\usepackage{lscape, adjustbox}
\usepackage{varwidth}
\usepackage{soul, xcolor, colortbl}
\usepackage{stfloats}
\usepackage{float}
\newcommand{\IGNORE}[1]{}
\usepackage{multirow}
\usepackage{xspace}
\newcommand{\dataset}{\texttt{MoVerb}\xspace}

\newcommand{\categoryone}{Quirk's categories\xspace}
\newcommand{\categorytwo}{Palmer's categories\xspace}

\newcommand{\moverbquirk}{\texttt{MoVerb}-Quirk\xspace}
\newcommand{\moverbpalmer}{\texttt{MoVerb}-Palmer\xspace}

\newcommand{\ruppenhofer}{R\&R\xspace}
\newcolumntype{P}[1]{>{\centering\arraybackslash}p{#1}}

\usepackage{makecell}
\usepackage{tabu, booktabs}
\usepackage{caption}
\usepackage{subcaption}
\usepackage{placeins}
\usepackage{enumitem}

\newcommand{\textbfsc}[1]{\textbf{\textsc{#1}}}

\newcommand{\smtextbfsc}[1]{\fontsize{7.7pt}{0.3cm}\textbf{\textsc{#1}}}

\usepackage{footnote}
\makesavenoteenv{tabular}
\makesavenoteenv{table}

\usepackage{color, colortbl}
\definecolor{Gray}{gray}{0.93}

\title{
Quirk or Palmer: A Comparative Study of Modal Verb Frameworks \\with Annotated Datasets}

\author{Risako Owan, \space Maria Gini, \space Dongyeop Kang \\
Computer Science and Engineering\\
University of Minnesota \\
\texttt{\{owan0002,gini,dongyeop\}@umn.edu} 
}

\begin{document}
\maketitle

\input{0-abstract}

\input{1-intro}

\input{2-related-work}

\section{Case Studies of Potential Applications}\label{sec:case_study}

There is some debate as to whether we should focus on modality as a whole since it can be expressed with other vocabularies not limited to modal verbs \cite{Nissim2013CrosslinguisticAO, Pyatkin2021ThePT}. However, we argue that modal verbs alone offer enough complexity, and there are still downstream NLP tasks that would benefit from better modal verb categorization. 

Difficulty with modal verb understanding can cause confusion in semantic similarity tasks. Using a RoBERTa Hugging Face model \cite{huggingface-transformer} pretrained on the Microsoft Research Paraphrase Corpus (MRPC) subset of the General Language Understanding Evaluation (GLUE) dataset\footnote{textattack/roberta-base-MRPC},  we saw that given some original sentence, the model was not able to reliably identify the unlikely interpretations. For example, given the sentence, ``My parents said I \textit{can} go'', the model would flag all following three as semantically equivalent by a score of at least 0.73: ``My parents said I have the ability to go.', ``My parents said I might go.'', and ``My parents said I have permission to go''.\footnote{0.978, 0.732, and 0.988 respectively}

As another example, we generated paraphrases for the Empathetic Dialogues dataset \cite{empathetic_dialogues} using the T5 Parrot paraphraser \cite{parrot} in the Hugging Face library.\footnote{prithivida/parrot\_paraphraser\_on\_T5} This revealed that $1951$ out of 2490 (78.35\%) paraphrases created for 865 sentences\footnote{We removed utterances with multiple sentences since paraphrase models will sometimes drop a sentence in an attempt to create a "new" paraphrase.} kept their original modal verbs. This shows that being able to correctly identify and paraphrase the sense of a modal verb can greatly increase variety in paraphrasing. 
\input{4-theoretical}

\input{5-dataset}

\input{6-evaluation}
\input{7-conclusion}

\bibliographystyle{acl_natbib}
\bibliography{acl2022}

\include{appendix}

\end{document}

%% file: 0-abstract.tex
\begin{abstract}
Modal verbs, such as \textit{can}, \textit{may}, and \textit{must}, are commonly used in daily communication to convey the speaker's perspective related to the likelihood and/or mode of the proposition. 
They can differ greatly in meaning depending on how they're used and the context of a sentence (e.g. ``They \textit{must} help each other out.'' vs. ``They \textit{must} have helped each other out.'')
Despite their practical importance in natural language understanding, linguists have yet to agree on a single, prominent framework for the categorization of modal verb senses. 
This lack of agreement stems from high degrees of flexibility and polysemy from the modal verbs, making it more difficult for researchers to incorporate insights from this family of words into their work.
This work presents \dataset dataset, which consists of 27,240 annotations of modal verb senses over 4,540 utterances containing one or more sentences from social conversations. Each utterance is annotated by three annotators using two different theoretical frameworks (i.e., Quirk and Palmer) of modal verb senses. 
We observe that both frameworks have similar inter-annotator agreements, despite having different numbers of sense types (8 for Quirk and 3 for Palmer). 
With the RoBERTa-based classifiers fine-tuned on \dataset, we achieve F1 scores of 82.2 and 78.3 on Quirk and Palmer, respectively, showing that modal verb sense disambiguation is not a trivial task.
Our dataset will be publicly available with our final version.\footnote{\url{https://github.com/minnesotanlp/moverb}}
\end{abstract}

%% file: 1-intro.tex
\section{Introduction}
Modal verbs (also referred to as modal operators, modals, or modal auxiliaries \cite{ImreAttila2017ALAt}) convey important semantic information about a situation being described or the speaker's perspective related to the likelihood and/or mode of the proposition \cite{Lyons1997, Quirk1985}.
Because of the widespread use of modal verbs in our daily lives, an accurate modeling of modal verb senses from context is essential for semantic understanding. For example, as modal verbs are often used with verbs that express one's personal state or stance, such as \textit{admit, imagine}, and \textit{resist} \cite{Biber}, we can utilize them for better speaker intention identification or sentiment analysis (See our case studies in Section \ref{sec:case_study}).

In both linguistics and NLP, however, there is no unifying consensus on how to organize these words (Table \ref{tab:framework_comparison}). One reason for this indeterminacy is their lack of a straightforward definition \cite{Nuyts2005OnDM}. 
Modal verbs have nuanced meanings, and their interpretation is often subjective.
If a speaker says, ``I \textit{can} go to the event today'', it can refer to their ability to go to the event.
Alternatively, if the speaker is a minor, the listener may interpret it as having permission from their parents. 
As such, categorizing modal verbs requires more attention than many other linguistic features, making the task challenging even for humans (Table \ref{tab:annotation_examples} shows examples of annotator disagreement on the modal verb categorization task).

We present a new dataset, \dataset, containing 4540 annotated conversational English utterances with their modal verb categories. We chose the conversational domain since spoken, casual text is more flexible and nuanced compared to language from other domains, and therefore could reap the most benefits from better modal verb classifications. 
To the best of our knowledge, this study provides the first empirical comparison of two modal verb frameworks with annotated datasets, evaluating the practicality of these different theoretical frameworks.
Our study shows a clear inclination towards one of the two frameworks and quantitatively shows how humans struggle with the task.

\begin{savenotes}
\begin{table*}[]
    \footnotesize
    \centering
    \begin{tabular}{@{}P{2.6cm}P{0.6cm}P{0.6cm}P{0.6cm}P{0.6cm}P{0.6cm}P{0.6cm}}
    \toprule
    \textbfsc{Reference} & \multicolumn{6}{c}{\textbfsc{Modality Categories}}\\
    \midrule
    \citet{Kratzer1991} &  \multicolumn{6}{c}{Epistemic{\hskip 3cm}Deontic{\hskip 3cm}Circumstantial}\\\midrule
    \citet{palmer_1986} & \multicolumn{6}{c}{Epistemic{\hskip 3cm}Deontic{\hskip 3cm}Dynamic}\\\midrule
    \citet{Quirk1985} & \multicolumn{6}{c}{Possibility{\hskip 3mm}Ability{\hskip 3mm}Permission{\hskip 3mm}Necessity\footnote{Logical Necessity}{\hskip 3mm}Obligation\footnote{Obligation/Compulsion}{\hskip 3mm}Inference\footnote{Tentative Inference}{\hskip 3mm}Prediction{\hskip 3mm}Volition}\\\midrule
    \citet{baker} & \multicolumn{6}{c}{Requirement{\hskip 5mm}Permissive{\hskip 5mm}Success{\hskip 5mm}Effort{\hskip 5mm}Intention{\hskip 5mm}Ability{\hskip 5mm}Want{\hskip 5mm}Belief}\\\midrule
    \citet{Ruppenhofer} & \multicolumn{6}{c}{Epistemic{\hskip 8mm}Deontic{\hskip 8mm}Dynamic{\hskip 8mm}Optative{\hskip 8mm}Concessive{\hskip 8mm}Conditional}\\\midrule
    \vspace{-4mm}\citet{Matthewson2018ModalFF} & \multicolumn{6}{c}{\makecell{Root \\ {(Teleological\hspace{0.2cm}Deontic\hspace{0.2cm}Bouletic)} }\hspace{0.2cm}\makecell{Epistemic\\ (Inferential\hspace{0.2cm}Reportative)}}\\\midrule
    \citet{Nissim2013CrosslinguisticAO}\footnote{\citeauthor{Nissim2013CrosslinguisticAO}'s work includes more categories on different dimensions, but we only show those comparable to the others in this table} & \multicolumn{4}{c}{\makecell{Epistemic\\(committment\hspace{0.2cm}evidential)}{\hskip 5mm}\makecell{Deontic\\(manipulative\hspace{0.2cm}volition)}} & \multicolumn{2}{c}{\makecell{Dynamic\\(axiological\hspace{0.2cm}appreciative\hspace{0.2cm}apprehensional)}}\\\midrule
    \citet{Portner} & \multicolumn{2}{c}{\makecell{Epistemic\\}} & \multicolumn{2}{c}{\makecell{Priority\\(Deontic \hspace{0.2cm} Bouletic \hspace{0.2cm} Teleological)}} & \multicolumn{2}{c}{\makecell{Dynamic\\(Volitional \hspace{0.2cm} Quantificational)}}\\
    \bottomrule
    \end{tabular}
    \caption{A non-exhaustive list of past work on modality and the frameworks they use. Note that some linguists support two-tiered categorical frameworks by defining general categories that are further divided into subcategories.}
    \label{tab:framework_comparison}
\end{table*}
\end{savenotes}

Our main contributions are as follows:
\begin{itemize}\setlength{\parskip}{0pt}\setlength{\itemsep}{0pt}\setlength{\partopsep}{0pt}
\item We collected \dataset, an annotated conversational domain dataset containing two types of labels for modal verbs in 4540 English utterances. The dataset is split into two: the first consisting of utterances with a single final label determined by majority voting and the second consisting of utterances with complete disagreement (Table \ref{tab:annotation_examples}).
\item We found the difficulty of annotating modal verbs even based on solid theoretical frameworks. We show that high annotator disagreement can be caused by issues not necessarily related to a difference in interpretation. We suggest that the gap between practice and theory could be bridged by prioritizing the straightforwardness of annotations.

\item We found a clear performance gap between the fine-tuned classifiers trained on different frameworks of data in \dataset: 82.2 F1 on Quirk and 78.3 F1 on Palmer.
Additionally, both frameworks struggled when applied across different domains.
\end{itemize}

%% file: 2-related-work.tex
\section{Related work}\label{sec:related_work}
\begin{table*}[t!] 
\small
\centering
\begin{tabular}{V{8.5cm}ccc}
\toprule
\textbfsc{Utterances with complete agreement}  & \textbfsc{Annotator 1} & \textbfsc{Annotator 2} & \textbfsc{Annotator 3}\\
\midrule
Usually moving your body helps but it depends on her situation... i \textit{would} get a 2nd opinion! & volition & volition & volition\\
\midrule
I bought a lottery ticket and have a feeling I \textit{will} win. & prediction & prediction & prediction\\
\midrule
That is really sweet of them. \textit{Must} have been a big party. & necessity & necessity & necessity\\
\midrule
I get it.. but you know life really is too short.. i think you \textit{should} try to reach out! Do it!:) & obligation & obligation & obligation\\
\midrule
\multicolumn{4}{c}{}
\\
\toprule
\textbfsc{Utterances with complete disagreement}  & \textbfsc{Annotator 1} & \textbfsc{Annotator 2} & \textbfsc{Annotator 3}\\
\midrule
That \textit{must} have been terrible. Were you okay? & inference & necessity & possibility\\
\midrule
I am going to a drink and paint party tomorrow. It \textit{should} be pretty fun! & inference & necessity & prediction\\
\midrule
I am stressed by my blood test results that I \textit{will} have tomorrow. & ability & necessity & prediction\\
\midrule
I work remotely, I wish that you \textit{could} do something like that as well.& ability & permission & possibility\\
\bottomrule
\end{tabular}
\caption{Annotation examples from \dataset for complete agreement and disagreement among the three annotators. Note that \textit{necessity} here refers to logical necessity, not social or physical necessities.}
\label{tab:annotation_examples}
\end{table*}

There are numerous linguistic studies about modal verbs and their categorization theories \cite{Quirk1985, palmerModality1990, Lyons1997, Mindt, kratzer2012, morante-sporleder-2012-modality, aarts_mcmahon_hinrichs_2021}. However, despite attempts to reconcile them \cite{reconcile_frameworks}, the widespread variation makes it unclear which theory and framework would work best for certain NLP tasks (Table \ref{tab:framework_comparison}). A dataset using multiple modal verb frameworks would help researchers experiment, but that dataset is yet to be built.
To the best of our knowledge, there is no English dataset dedicated to the comparison of modal verb labeling.

Framework consistency is not the only thing lacking in modality datasets. 
Sources of modality can vary from dataset to dataset as well. In a multilingual corpus focusing on modality as a whole, \citeauthor{Nissim2013CrosslinguisticAO} manually tag words and phrases representing modality. Because of the lack of emphasis on modal verbs, this dataset contains only 32 instances over 7 modal verbs: \textit{will, might, can, may, would, could,} and \textit{should} \cite{Nissim2013CrosslinguisticAO}. Although some may argue that focusing on modality as a whole is more natural, we argue that a dataset focusing on modal verbs is also necessary because of the ample complexities of modal verbs on their own, as previously mentioned.

Even datasets that do focus on modal verbs are not guaranteed to study the same set of words \cite{Ruppenhofer, marasovic-etal-2016-modal}. Additionally, modal verbs in different domains, namely conversational and academic, have quite dissimilar distributions \cite{Biber}. 
In our cross-domain analysis, we utilized a dataset for subjectivity analysis in opinions and speculations from the news domain \cite{Ruppenhofer, Wiebe2005AnnotatingEO}. 
\citeauthor{Ruppenhofer} do not include \textit{would} and \textit{will} in their annotations, making their dataset challenging for analyzing conversational English. \textit{Would} and \textit{will} are 1st and 3rd when we rank modal verbs by their frequencies in spoken English \cite{Mindt, Biber}.\footnote{As of August 2022, the Corpus of Contemporary American English (COCA) dataset \url{https://www.wordfrequency.info} shows \textit{will} and \textit{would} being the 47th and 49th most common words overall. For comparison, \textit{shall} is 1090th.} 

We note that there is a slight difference in our annotation frameworks. \citeauthor{Ruppenhofer} create a different schema of their own, building off of work by \citet{baker} and \citet{palmer_1986}. We do not use \citeauthor{baker}'s labels since we are more interested in applying traditional linguistic theories. However, we are still able to compare results since \categorytwo make up 97.57\% of the annotations in the \citeauthor{Ruppenhofer} dataset.

%% file: 4-theoretical.tex
\section{Theoretical Frameworks}
\begin{table*}
\small
\centering
\begin{tabular}{l c c c ccc cc}
\toprule
&\smtextbfsc{possibility}&{\smtextbfsc{prediction}}&\smtextbfsc{inference}&\smtextbfsc{necessity}&\smtextbfsc{ability}&\smtextbfsc{volition}&\smtextbfsc{permission}&\smtextbfsc{obligation}\\
\midrule
\smtextbfsc{deontic} & 50 & 21 & 22 & 27 & 42 & 31 & 22 & 288\\
\smtextbfsc{epistemic} & 454 & 307 & 120 & 317 & 110 & 12 & 1 & 10\\
\smtextbfsc{dynamic} & 197 & 172 & 13 & 11 & 758 & 194 & 22 & 22\\
\bottomrule
\end{tabular}
\caption{The frequency distribution between Quirk's and \categorytwo in \dataset. This table shows that there is no clear mapping between the two frameworks, although there are common combinations (epistemic possibility, dynamic ability, etc.) that reveal overlapping categories. }
\label{tab:7_3_table}
\end{table*}

We use two labeling frameworks in our dataset annotations that we refer to as \categoryone and \categorytwo. The labels we use are as follows:

\vspace{-3mm}
\begin{itemize}\setlength{\partopsep}{0pt} \setlength{\itemsep}{0pt}\setlength{\parskip}{0pt}
    \item \categoryone consist of eight labels: \textit{possibility},
    \textit{ability},
    \textit{permission},
    \textit{logical necessity} (abbrev. \textit{necessity}),
    \textit{obligation/compulsion} (abbrev. \textit{obligation}),
    \textit{tentative inference} (abbrev. \textit{inference}),
    \textit{prediction}, and \textit{volition}. While each category's definitions may be inferred by their names, further explanations can be found in Figures \ref{fig:mturk_quirk_descriptions} and \ref{fig:mturk_quirk_examples} in Appendix \ref{sec:mturk_form}.
    \item \categorytwo consist of three labels: \textit{deontic}, \textit{epistemic}, and \textit{dynamic}. A deontic modal verb influences a thought, action, or event by giving permission, expressing an obligation, or making a promise or threat. An epistemic one is concerned with matters of knowledge or belief and with the possibility of whether or not something is true. Lastly, dynamic modal verbs are related to the volition or ability of the speaker or subject, in other words, some circumstantial possibility involving an individual (Figures \ref{fig:mturk_palmer_descriptions} and \ref{fig:mturk_palmer_examples} in Appendix \ref{sec:mturk_form}).
\end{itemize}

Table \ref{tab:7_3_table} shows a contingency table for \dataset. We see that there is no straightforward mapping allowing us to cleanly convert one framework to the other. However, the different distributions of one set of labels within labels of the other framework reveal which categories are similar to each other.

%% file: 5-dataset.tex
\section{\dataset: Annotated Modal Verb Dataset}\label{sec:methodology}

\begin{table*}[t]
\small
\centering
\begin{tabular}{l@{\hskip 1mm}c@{\hskip 2mm}c@{\hskip 2mm}c@{\hskip 2mm}c@{\hskip 2mm}c@{\hskip 2mm}c@{\hskip 2mm}c@{\hskip 2mm}c@{\hskip 2mm}r}
\toprule
&\textbfsc{will}&\textbfsc{would}&\textbfsc{should}&\textbfsc{may}&\textbfsc{might}&\textbfsc{must}&\textbfsc{could}&\textbfsc{can}&\textbfsc{total}\\
\midrule
\smtextbfsc{possibility} & 50 & 61 & 7 & 128 & 324 & 0 & 119 & 96 & 785 (0.22\%)\\
\smtextbfsc{ability} & 14 & 24 & 0 & 0 & 0 & 1 & 302 & 657 & 998 (0.28\%)\\
\smtextbfsc{permission} & 2 & 4 & 4 & 19 & 1 & 0 & 10 & 12 & 52 (0.01\%)\\
\smtextbfsc{necessity} & 7 & 12 & 13 & 0 & 0 & 334 & 3 & 1 & 370 (0.1\%)\\
\smtextbfsc{obligation} & 5 & 6 & 307 & 1 & 0 & 18 & 0 & 4 & 341 (0.1\%)\\
\smtextbfsc{inference} & 6 & 42 & 45 & 2 & 11 & 73 & 1 & 1 & 181 (0.05\%)\\
\smtextbfsc{prediction} & 351 & 183 & 19 & 0 & 5 & 4 & 4 & 3 & 569 (0.16\%)\\
\smtextbfsc{volition} & 129 & 92 & 11 & 3 & 6 & 1 & 6 & 6 & 254 (0.07\%)\\
\midrule
\textbfsc{total} & 564 (16\%) & 424 (12\%) & 406 (11\%) & 153 (4\%) & 347 (10\%) & 431 (12\%) & 445 (13\%) & 780 (22\%) & \textbf{3550}\\
\toprule
\smtextbfsc{epistemic} & 283 & 269 & 78 & 99 & 232 & 479 & 118 & 161 & 1719 (42\%)\\
\smtextbfsc{deontic} & 32 & 65 & 437 & 25 & 18 & 35 & 27 & 52 & 691 (16.9\%)\\
\smtextbfsc{dynamic} & 336 & 258 & 29 & 37 & 108 & 6 & 315 & 592 & 1681 (41.1\%)\\
\midrule
\textbfsc{total} & 651 (16\%) & 592 (14\%) & 544 (13\%) & 161 (4\%) &358 (9\%) &520 (13\%) &460 (11\%) &805 (20\%) & \textbf{4091}\\
\bottomrule
\end{tabular}
\captionof{table}{The breakdown of categories in each modal verb.}
\label{tab:dataset-statistics}
\end{table*}

Table \ref{tab:dataset-statistics} shows the  statistics of our \dataset dataset.
We use the eight core modal verbs in our study:  \textit{can}, \textit{could}, \textit{may}, \textit{might}, \textit{must}, \textit{will}, \textit{would}, and \textit{should}.
\textit{Shall} is also another core modal verb but is excluded from our work since there are too few instances of it in our conversational dataset.\footnote{\textit{shall} is more likely to be used in legal contexts \cite{CoatesAndLeech}, which is outside the scope of this study.}

We chose the Empathetic Dialogues dataset \cite{empathetic_dialogues} for our annotation task because of its variety of utterances in the conversational domain and wide usage in social dialogues. An utterance is defined as a speaker's output in a single turn and can potentially be one or more sentences. We extracted utterances containing only one modal verb as detected using SpaCy's POS tagger and lemmatizer \cite{spacy2}. We focused on utterances containing one modal verb for simplicity, but this did not exclude much from our dataset since only 2.4\% of the utterances had more than one modal verb.\footnote{78.8\% had none and 18.8\% had one.}
 
\begin{figure*}
    \begin{subfigure}[t]{0.32\textwidth}
        \centering
    \includegraphics[width=\textwidth,trim={0 0 0 0cm},clip]{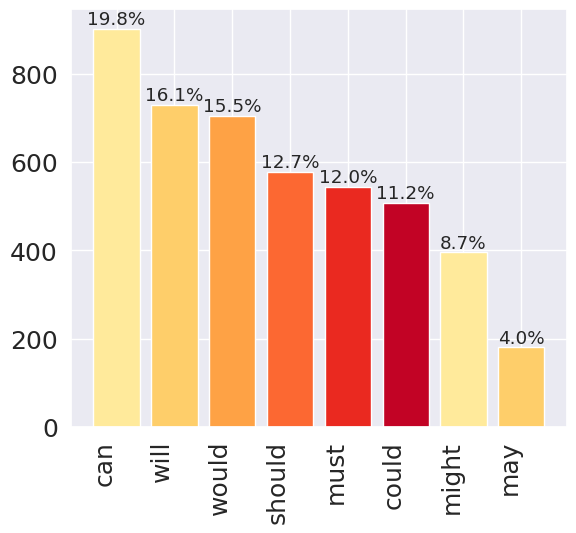}
        \caption{Modal verb distribution} 
        \label{fig:bar-a}
    \end{subfigure}\hfill
    \begin{subfigure}[t]{0.30\textwidth}
        \includegraphics[width=\textwidth]{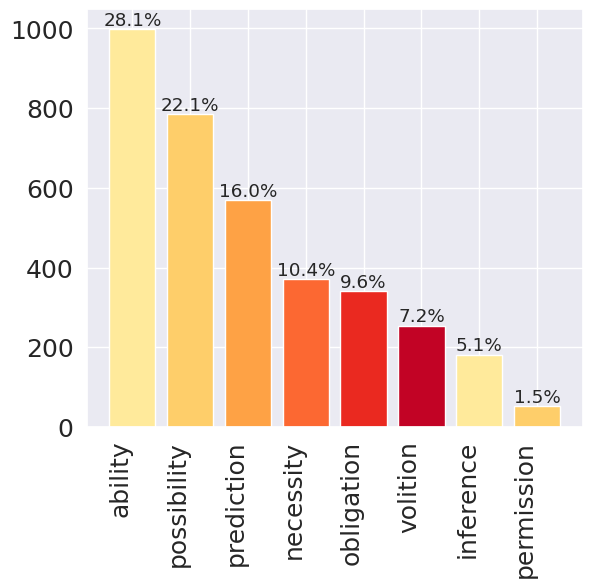}
        \caption{Quirk's label distribution. }
        \label{fig:bar-b}
    \end{subfigure}\hfill
    \begin{subfigure}[t]{0.30\textwidth}
        \includegraphics[width=\textwidth]{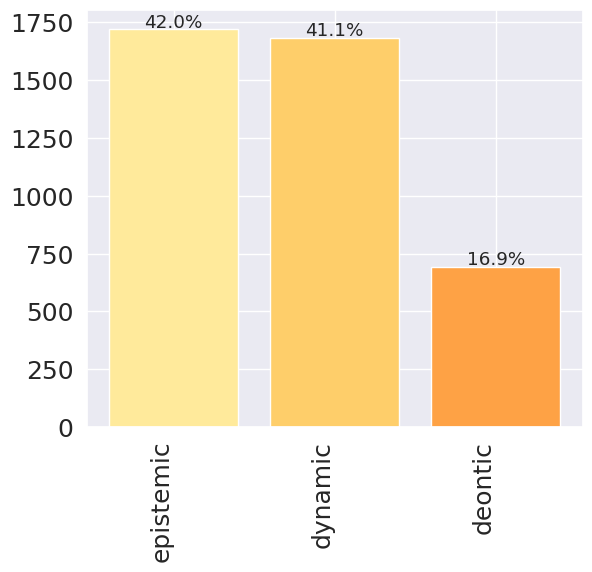}
        \caption{Palmer's label distribution.  }
        \label{fig:bar-c}
    \end{subfigure}
    \caption{Dataset statistics: (a) Modal verb distribution, (b) Quirk's categories label distribution, and (c) Palmer's categories label distribution. 
    This chart only includes utterances that had a majority label.}
    \label{fig:bar}
\end{figure*}

We included utterances containing more than one sentence (as long as they used only one modal verb) in order to retain as much context as possible. In this way, we separated out the first 4540 utterances containing single modal verbs, except for \textbf{may} and \textbf{might}, which we collected and used all of due to scarcity (Figure \ref{fig:bar-a} and Table \ref{tab:dataset-statistics}).

After finalizing our candidate utterances to annotate, we utilized Amazon Mechanical Turk (MTurk) to gather crowd-sourced labels for each modal verb. Three annotations were collected for each of the 4540 utterances and assigned final labels based on majority voting (Figures \ref{fig:bar-b} and \ref{fig:bar-c}). Our HIT (Human Intelligence Tasks) form is included in Appendix \ref{sec:mturk_form} (Figures \ref{fig:mturk_instructions}-\ref{fig:mturk_sentences_cat2}). 
We limited our MTurk pool to Master workers (high-performing workers) living in the US with approval rates of $>98\%$. 
Each worker was allowed to annotate as many HITs as they wanted 
and were allowed to submit annotations for both frameworks. They were prevented from participating any further if we saw that their annotations for \categoryone seemed random (Appendix \ref{sec:filtering_criteria}).
Unfortunately, we could not reliably differentiate randomness from subjective differences for \categorytwo, since each category has less stringent restrictions on which modal verbs can be assigned to them. As such, there is no clean and unsupervised method of quickly determining data quality for this framework.


\subsection{Post-analysis on Annotations}
\label{sec:post_analysis}


Our final annotations revealed some commonly disagreed upon categories (Figure \ref{fig:cat1_disagreements}, \ref{fig:cat2_disagreements} here and Table \ref{tab:conflicting_annots} in Appendix \ref{sec:dataset_statistics}). For some instances in \categoryone, it seemed that annotators were simply using certain labels interchangeably, as opposed to truly diverging on how the modal verb affected the utterance. For example, in Figure \ref{fig:cat1_disagreements}, we can see that \textit{inference} and \textit{(logical) necessity} are co-occurring in high frequencies. Utterances containing sentences like, ``You \textit{must} have been so happy'' and ``You \textit{must} have been so scared'' in the dataset often both had at least one \textit{(logical) necessity} and \textit{inference}  annotations each. While sentence length is not an infallible measure of context, the lack of correlation between it and annotator disagreement (Figure \ref{fig:context_effect} in Appendix \ref{sec:dataset_statistics}) suggests that simply having more text may not be the best method for providing context for this task.
This illustrates how frameworks well-grounded in theory can still be interpreted differently in practice. Some labels may be combined for less noise if their distinction is unnecessary for downstream tasks.

\begin{table}[t]
\small
\centering
\begin{tabular}{lcc}
\toprule
&\textbfsc{Quirk} & \textbfsc{Palmer}\\
\midrule
\textbfsc{Percent agreement} &0.58 & 0.60\\
\bottomrule
\end{tabular}
\caption{Percent agreement values for both frameworks in \dataset}
\vspace{-3mm}
\label{tab:percent_agreement}
\end{table}

Another common behavior was that annotators sometimes seemed to label utterances based on what could be inferred. For example, an utterance containing a sentence like ``I \textit{may} go to the store today'' was often labeled as both \textit{ability} and \textit{possibility}. One could argue that this \textit{may} represents \textit{ability}, since it indicates that the user has the ability to go to the store today or that the information regarding the speaker's \textit{ability} is most important. However, one could also claim that the annotator is then labeling what can be inferred from the utterance, not necessarily what the modal verb semantically represents. This behavior could also be observed in Table \ref{tab:7_3_table} and Figure \ref{fig:cat2_disagreements}, where \textit{epistemic} and \textit{dynamic}, which are seen most in utterances labelled \textit{possibility} and \textit{ability}, appear commonly in conflicting annotation pairs.

All in all, annotators seemed to struggle more with using Palmer's categories, as can be speculated from the fact that the percent agreement between the two frameworks was very similar, despite Palmer's categories having significantly fewer labels (Table \ref{tab:percent_agreement}). This is not too surprising, given that Palmer's categories are more abstract and can thus be less intuitive. The unfamiliar label titles may have also added a layer of complication to the task.


\subsection{Data Subjectivity}
\label{sec:data_subjectivity} 

We argue that these disagreements highlight the flexibility and ambiguity that have plagued linguists for decades and emphasize the subjectivity of modal verbs. This is because modal verb annotations highly rely on what the reader interprets as the main takeaway of what the modal verb represents. 
Quirk's mappings (Table \ref{tab:quirk_categories} in Appendix \ref{sec:dataset_statistics}) were not used to limit annotator options in the MTurk form since we wanted annotators to select labels on their own with minimal input from us. The added flexibility may have led to lower inter-annotator agreement levels, which is inevitable for subjective annotations \cite{Leonardelli2021AgreeingTD, Basile2020ItsTE, Aroyo}.

%% file: 6-evaluation.tex
\begin{figure}[t]
    \centering
     \hspace*{-0.5cm}\includegraphics[scale=0.68,trim={0.2cm 0 0 0}, clip]{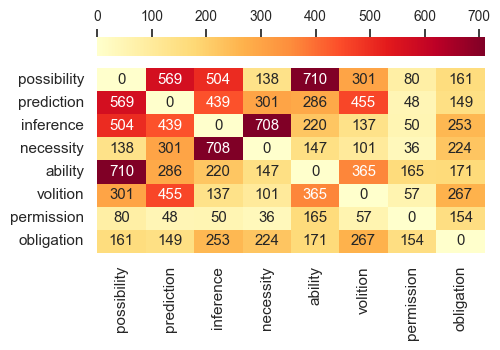}\vspace{-4mm}
    \caption{The frequency of disagreeing annotation pairs in \categoryone. By disagreement, we mean when two annotators do not choose the same label for some given utterance. Each utterance can have 3 counts of disagreements because there are 3 possible annotation pairs. }
    \label{fig:cat1_disagreements}
\end{figure}

\begin{figure}[t]
    \centering
    \includegraphics[scale=0.62,trim={0.2cm 0 0 1.7cm}, clip]{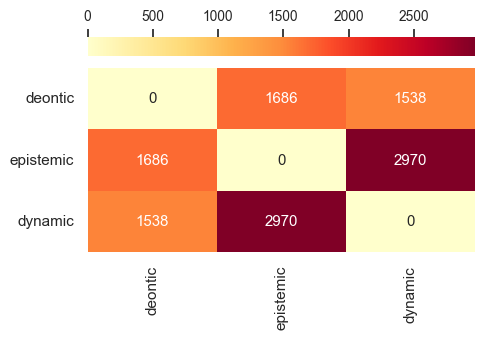}\vspace{-4mm}
    \caption{The frequency of disagreeing annotations in \moverbpalmer. This uses the same logic as Figure \ref{fig:cat1_disagreements}}
    \label{fig:cat2_disagreements}
\end{figure}

\section{Experimental Results}
We like to answer the following questions using the collect \dataset dataset: (1) how well \dataset could be used to train the latest deep learning models for modal verb sense prediction task (Section \ref{sec:single_domain_classification}) and (2) how transferable that knowledge (trained on the conversational domain) was to other domains, namely the news opinion domain (Section \ref{sec:cross_domain_transferability}). 

\paragraph{Experiment Design}\label{sec:evaluation}
For the first experiment, we split our datasets into train-test ratios of 90/10. For the second experiment focusing on transferability, we used one dataset for the training and validation data and the other for the test set. We conducted this on both \moverbpalmer$\rightarrow$ \citeauthor{Ruppenhofer} and \citeauthor{Ruppenhofer}$\rightarrow$ \moverbpalmer combinations. Additionally, since we initially surmised that the lack of \textit{will/would} examples in the \citeauthor{Ruppenhofer} dataset would cause issues, we conducted the same experiment with those modal verbs removed from \moverbpalmer to observe the effect of not including \textit{will/would} (Table \ref{tab:model_comparison} and \ref{tab:x_domain_transferability}).

For all experiments, we ran 10-fold cross-validations and used an early stopping callback that would get triggered once the F1 value stopped increasing by at least 0.01.
For learning rates, we tested among $5e-6, 1e-5$, and $2e-5$, and used the weighted F1 score for evaluation. We used the Pytorch Lightning library to train and evaluate a Transformer model with an Adam epsilon of 1e-8, and a batch size of 32. Additionally, our trainer used GPU acceleration with a GeForce RTX 3090 using the DistributedDataParallel strategy.

We fine-tuned six Transformer-based models \cite{Vaswani} from Huggingface Transformers \cite{huggingface-transformer}: ALBERT\textsubscript{base} \cite{albert}, BERT (both base and large) \cite{Devlin2019BERTPO}, RoBERTa (both base and large) \cite{liu2019roberta}, and DistilBERT\textsubscript{base} \cite{distilbert}.
In all runs, the RoBERTa models showed the best test F1 scores (Tables \ref{tab:classifier_res2} and \ref{tab:classifier_res4} in Appendix \ref{sec:classification_results}).

\begin{table}[t]
\small
\centering
\begin{tabular}{lcc}
\toprule
\textbfsc{Dataset} & \textbfsc{Validation} & \textbfsc{Test}\\ 
\midrule
\moverbquirk & 78.98 & 82.22\\ 
\moverbquirk (w/o will/would) & 83.56 & 84.31\\
\midrule
\moverbpalmer & 77.08 & 78.36\\ 
\moverbpalmer (w/o will/would) &  80.62 & 80.89\\ 
\midrule
\citeauthor{Ruppenhofer} & 83.31 & 85.60\\ 
\hline
\end{tabular}
\caption{Best-performing F1 scores averaged over a 10-fold cross validation traned on various datasets. We selected the best F1 scores out of various model and learning rate combinations. For a more complete table, see Table \ref{tab:classifier_res2}.}
\label{tab:model_comparison}
\end{table}

\begin{table*}[t] 
\small
\centering
\begin{tabular}{ccccc}
\toprule
\textbfsc{Rank} & \multicolumn{2}{c}{\textbfsc{\moverbpalmer}} & \multicolumn{2}{c}{\textbfsc{\citeauthor{Ruppenhofer}}}\\
& \smtextbfsc{Modal Verb} & \smtextbfsc{Label} & \smtextbfsc{Modal Verb} & \smtextbfsc{Label}\\
\midrule
1 & can (19.7\%) & epistemic (42.0\%) & can (29.5\%) & deontic (46.1\%)\\
\midrule
2 & will (15.9\%) & dynamic (41.1\%) & should (22.4\%) & epistemic (27.6\%)\\
\midrule
3 & would (14.5\%) & deontic (16.9\%) & could (19.7\%) & dynamic (26.3\%)\\
\midrule
4 & should (13.3\%) & - & must (14.8\%) & -\\
\midrule
5 &  must (12.7\%) & - & may (8.5\%) & -\\
\bottomrule
\end{tabular}
\caption{Modal verb and label distribution comparisons between \dataset and \citet{Ruppenhofer}. Note that while the modal verb ranking will be the same for both frameworks in \dataset, we only list a ranking of \moverbpalmer in order to compare it with \citet{Ruppenhofer}.}
\label{tab:ruppenhofer_vs_palmer}
\end{table*}

\subsection{Single-Domain Modal Verb Sense Classification }\label{sec:single_domain_classification}

From Table \ref{tab:model_comparison}, we observe that \dataset can indeed be used to train Transformer-based models \cite{Vaswani} on how to label modal verbs.
The table shows that \moverbquirk does better at training models compared to \moverbpalmer.
However, we see that performance differences between domains have a larger effect than comparisons observed between frameworks. The performance difference between \moverbpalmer and \citeauthor{Ruppenhofer}'s dataset is greater than that between \moverbpalmer and \moverbquirk. This was even after removing \textit{will}s and \textit{would}s, since \citeauthor{Ruppenhofer} did not annotate those two modal verbs, which were common in our Disagreement subset. This greater performance difference may be attributed to the fact that news-related writing tends to be more structured than conversational data or that the \citeauthor{Ruppenhofer}'s dataset contained a higher proportion of \textit{should} and \textit{could}s, which were less likely to be disagreed upon (Table \ref{tab:ruppenhofer_vs_palmer} here and Table \ref{tab:agreement_disagreement_table_count} in Appendix \ref{sec:dataset_statistics}). However, more annotations in different domains should be collected for a further definitive conclusion.

Table \ref{tab:difficult_predictions} contains instances where the classifiers predicted incorrectly and with low confidence. Classification of these utterances is especially difficult because of the ambiguity of the modal verbs and need for more context. However, Table \ref{fig:context_effect} in Appendix \ref{sec:dataset_statistics} showed that text length was not correlated with better predictions in our dataset. This suggests that the context our model requires in \dataset may need to be collected from knowledge bases or by using long context comprehension.

\subsection{Cross-Domain Transferability}
\label{sec:cross_domain_transferability}
We also applied the classifiers trained on \moverbpalmer to the \citeauthor{Ruppenhofer} news opinion domain dataset\footnote{Downloaded from\\ \url{http://ruppenhofer.de/pages/Data\%20sets.html}} in order to see how our classification model might perform in another domain (Table \ref{tab:x_domain_transferability}).
As mentioned in Section \ref{sec:related_work}, this dataset uses a slightly modified framework, adding three more labels to Palmer's categories. However, we removed them in our experiment since they only made up 3.2\% of the dataset we extracted. We also filtered out sentences with more than one modal verb in order to mirror what we use in Empathetic Dialogues \cite{empathetic_dialogues}.

\begin{table}[h]
\small
\centering
\begin{tabular}{lll}
\toprule
\textbfsc{Dataset} & \textbfsc{Val. F1} & \textbfsc{Test F1}\\
\midrule
\moverbpalmer$\rightarrow$ \ruppenhofer & 75.4 & 61.44\\ 
\midrule
\ruppenhofer$\rightarrow$ \moverbpalmer & 86.5 & 66.37\\ 
\midrule
\moverbpalmer (w/o w$^2$)$\rightarrow$ \ruppenhofer & 80.23 & 69.74\\ 
\midrule
\ruppenhofer$\rightarrow$ \moverbpalmer (w/o w$^2$) & 86.5 & 75.93\\ 
\bottomrule
\end{tabular}
\caption{Observing cross-domain transferability. We use \ruppenhofer to represent \citeauthor{Ruppenhofer} and w$^2$ to represent \textit{will/would} in the interest of space. The dataset to the left of the arrow represents the training dataset, and the one on the right represents the validation/test set.}
\label{tab:x_domain_transferability}
\end{table}

We see that our models struggled significantly when the training data and test data came from different sources (Table \ref{tab:x_domain_transferability} here and Table \ref{tab:classifier_res4} in Appendix \ref{sec:dataset_statistics}). Utterances from a conversational dataset are bound to be different from opinions extracted from news sources due to the nature of their content. We additionally ran the same experiment after removing \textit{will/would} from \moverbpalmer to see the extent to which the lack of these two labels affected the F1 scores. The scores rose significantly for both directions although did not reach performance levels observed in single-domain classification (Table \ref{tab:x_domain_transferability}). A couple difficult examples for cross-domain classification is shown in Table \ref{tab:difficult_predictions} as well.



%% file: 7-conclusion.tex
\section{Conclusions}\label{sec:conclusion}

\begin{table*}[h!]
\centering
\small
\begin{tabular}{lV{6cm}lll}
\toprule
\textbfsc{Dataset} & \textbfsc{Sentence} & \textbfsc{Prediction} & \textbfsc{Label}\\
\midrule
\moverbquirk & We do not have a fence but I know my dog \textit{will} stay in the yard & volition (49.36) & prediction (3/3)\\
\moverbpalmer & That stinks! Try not to be jealous though. Something else \textit{will} come your way. & dynamic (49.87) & epistemic (3/3)\\
R\&R & ``A government in which the president controls the Supreme Court, the National Assembly and the Armed Forces \textit{can} not be called a democracy, '' Soto charged. & deontic (65.7) & dynamic (N/A)\\
\moverbpalmer$\rightarrow$ R\&R & They are provided with a medical exam upon admission, and their diet ( is someone making a point about diversity here ? ) ranges from bagels and cream cheese to rice and beans -- all eaten with plastic utensils -- after which the prisoners \textit{may} clean their teeth with specially shortened brushes. & epistemic (48.95) & deontic (N/A)\\
R\&R$\rightarrow$\moverbpalmer & News that big \textit{would} be a shock to anyone! How did you both handle it? & dynamic (49.74) & epistemic (3/3)\\
\bottomrule
\end{tabular}
\caption{Difficult examples incorrectly labeled by our RoBERTa-large classifier. The numbers in the parentheses represent the classifier's confidence score for the predictions and the annotator agreement score for the labels. We use R\&R to represent \citeauthor{Ruppenhofer}. Although coincidentally our \dataset instances here all have complete annotator agreement, they represent cases with ambiguous meaning and/or alternative interpretation. In the first example, we see that the model is focusing more on the dog by putting emphasis on its decision (volition) rather than its owner's prediction. In the second example, one could argue that the model ``believes'' more in the internal locus of control, where one's own actions determine an outcome (dynamic), as opposed to the external locus of control, which puts more emphasis on plain luck (epistemic). As such, instances with low prediction confidences can help shed light on alternative interpretations and hidden biases in text.}
\label{tab:difficult_predictions}
\end{table*}
We compared two linguistic frameworks by crowd-sourcing annotations for 4540 utterances. Modal verb categorization is a difficult task, even for humans, making supervised datasets a vital part of analyses. See Table \ref{tab:difficult_predictions} for examples our classifier struggled with. Our work shows that within \dataset, annotators had an easier time working with \categoryone. Fewer disagreement relative to the number of labels led to less noise, which translated to better performance on Transformer models, both intra and cross-domain, although domain differences seemed to affect the performance more than framework chosen (Table \ref{tab:classifier_res2} in Appendix \ref{sec:classification_results}). Additionally, \categoryone gave us a more precise study of modal verb patterns (Figure \ref{fig:cat1_disagreements} vs. Figure \ref{fig:cat2_disagreements}) due to more specific labels. Therefore, barring cases where there is a specific reason to use \categorytwo (i.e. expanding another dataset that uses \categorytwo or comparing work with other studies that use it), \textbf{we recommend working with \categoryone for smoother dataset generation and better downstream task performance}.

Our dataset will be available to the public and we hope that it will provide helpful information and insights for other studies as well.
Each framework's dataset will be split into two subsets: those with a label majority with at least 2 annotators agreeing with each other (Agreement subset) and those where there were absolutely no agreement among annotators (Disagreement subset\footnote{However, this disagreement subset is not used in our experiments.}) (Table \ref{tab:agreement_disagreement_table_count}).
Our fine-tuned classifiers will also be released with the dataset for those who need an easy-to-use modal verb intent classifier or find that it can help performance in other tasks when combined with other resources.

\section{Limitations and Future Work}\label{sec:discussion_and_future_work}
We list several limitations to our work and future plans for handling them. Firstly, this research does not consider modality in other languages, domains, or frameworks. Our conclusions and insights can only be applied to conversational instances of languages that share the same modal verb morphology as English. Therefore, expanding our target text and incorporating more frameworks can greatly increase potential uses for our dataset and thus would be an important next step.

Secondly, our data forces a single label onto each utterance. This is beneficial for training models, but could also mean we are disregarding disagreements that could shed more light onto how people interpret modal verbs. Methods of how to annotate subjective data and handle disagreement have been explored by many \cite{Basile2020ItsTE, Basile2019, Aroyo}. We believe our dataset can be used to test these strategies that propose modifications preventing disagreement to be treated as noise. Future work could also potentially include allowing annotators to express uncertainty on given labels.

The last limitation we discuss is that this work only focuses on utterances with single modal verbs. We would need to conduct more studies to determine how generalizable our work is to longer, more complicated sentences. This may involve working on another domain since conversational text is less often long and complicated. 

Other steps for advancing this work would be to use the dataset for specific NLP tasks, such as paraphrasing and inference. One way in which model verbs could be used in inference is to focus on \textit{permission} and \textit{obligation} to see social power dynamics in text (who seems to be receiving/giving permission more than average or who seems to be controlled by more social obligations). Or perhaps, one could investigate the annotations with complete disagreements and determine what caused those disagreements. Identifying what part of the sentence or context prompted certain annotations and lack of agreement would entail high degrees of natural language understanding.



\subsection*{Ethical Considerations}\label{sec:ethics}
We paid $\$1$ for 20 annotated sentences on MTurk, which translated to an average hourly wage of $\$12$. This is higher than both the federal and state minimum wage according to the Minnesota
Department of Labor and Industry. Additionally, recognizing the fact that our HITS were not easy and that annotator blocks can lead to terminated accounts, we utilized qualifications\footnote{Qualifications allow us to blacklist workers who did not reach our standards for this particular task, without jeopardizing their account status.} to prevent workers from submitting additional HITS to our project.
%

\subsection*{Acknowledgments}\label{sec:acknowledgments}
We thank Dr. Brian Reese and our anonymous reviewers for their valuable, constructive feedback. We would also like to extend our gratitude to the Mechanical Turk workers who made this project possible.

%% file: appendix.tex
\appendix
\begin{minipage}{1\textwidth}
\section{Experiment Design}
\subsection{Mechanical Turk}
\label{sec:mturk_form}
\vspace{0.5cm}
    \includegraphics[scale=0.45]{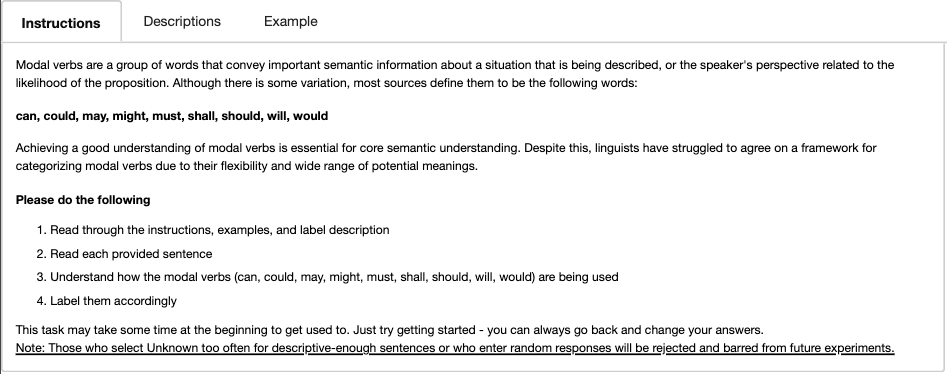}
    \captionof{figure}{General instructions given to MTurk workers}  
    \label{fig:mturk_instructions}
\vspace{1cm}
    \includegraphics[scale=0.45]{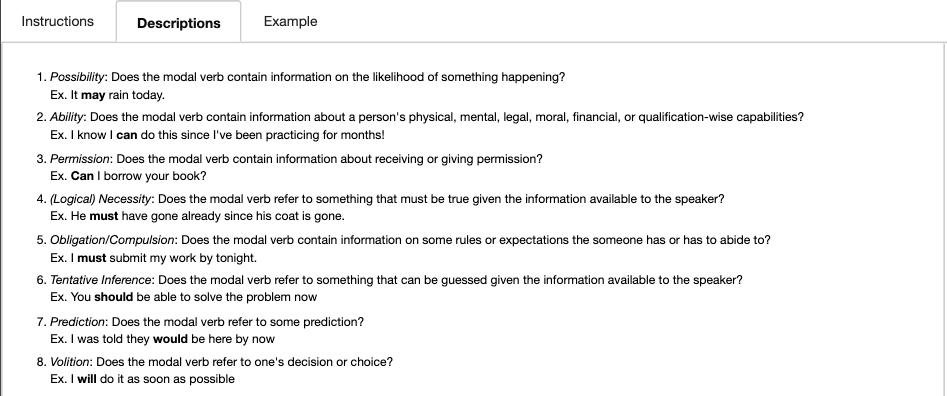}
    \captionof{figure}{Descriptions given to MTurk workers for \categoryone}
    \label{fig:mturk_quirk_descriptions}
\vspace{1cm}
    \includegraphics[scale=0.45]{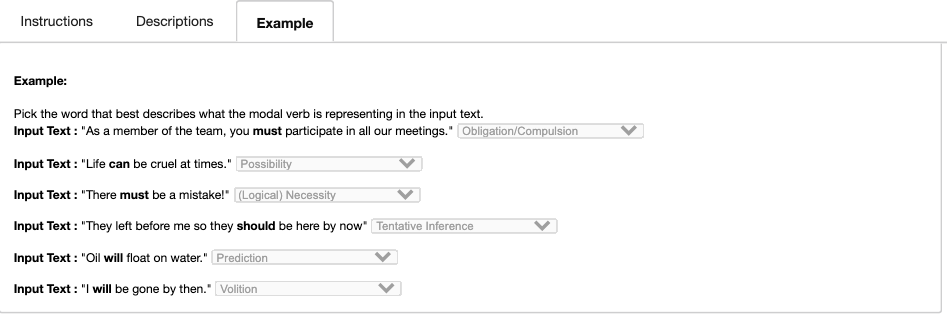}
    \captionof{figure}{Examples given to MTurk workers for \categoryone}  
    \label{fig:mturk_quirk_examples}
\end{minipage}

\clearpage

\begin{minipage}{1\textwidth}
    \includegraphics[scale=0.49]{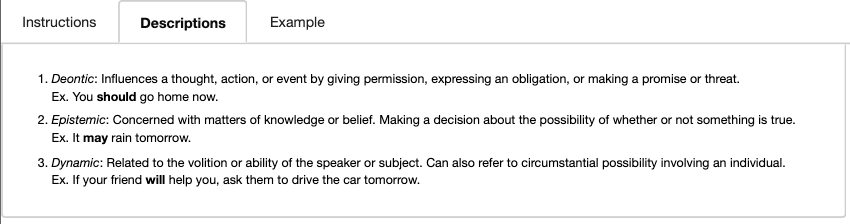}
    \captionof{figure}{Descriptions given to MTurk workers for \categorytwo}
    \label{fig:mturk_palmer_descriptions}
\vspace{1cm}
    \includegraphics[scale=0.49]{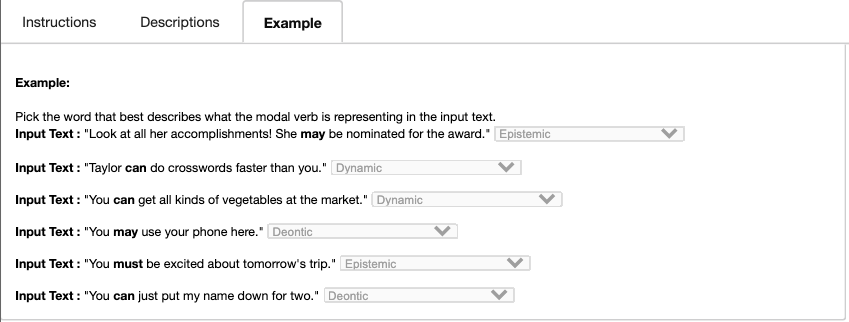}
    \captionof{figure}{Examples given to MTurk workers for \categorytwo}  
    \label{fig:mturk_palmer_examples}
\vspace{1cm}
    \includegraphics[scale=0.41]{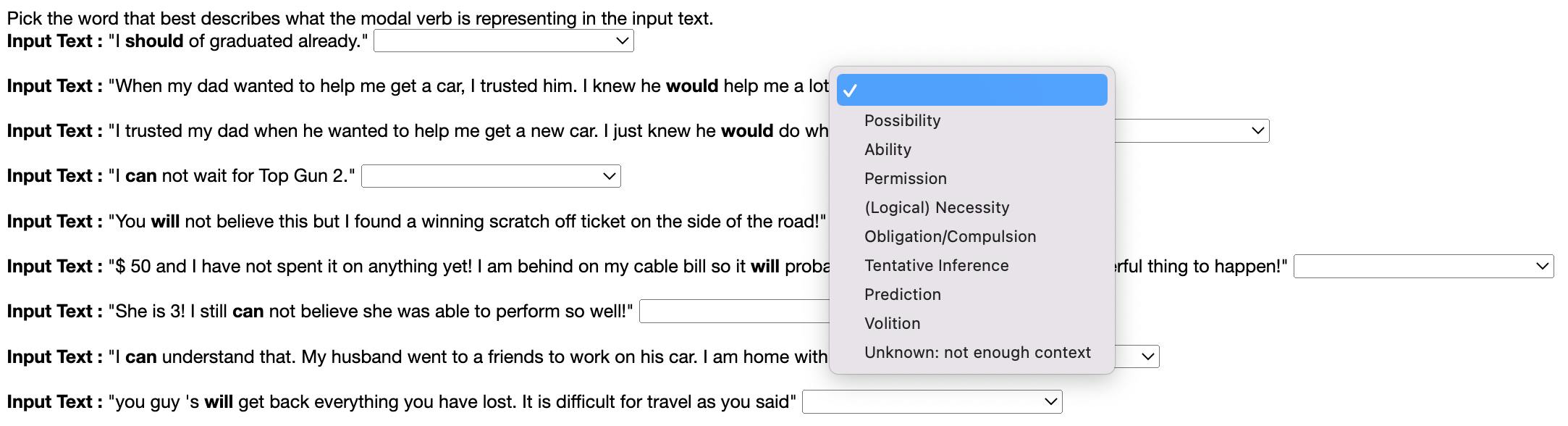}
    \captionof{figure}{Example sentences to annotate and the corresponding drop-down boxes for Quirk's categories}  
    \label{fig:mturk_sentences}
\vspace{1cm}
    \includegraphics[scale=0.41]{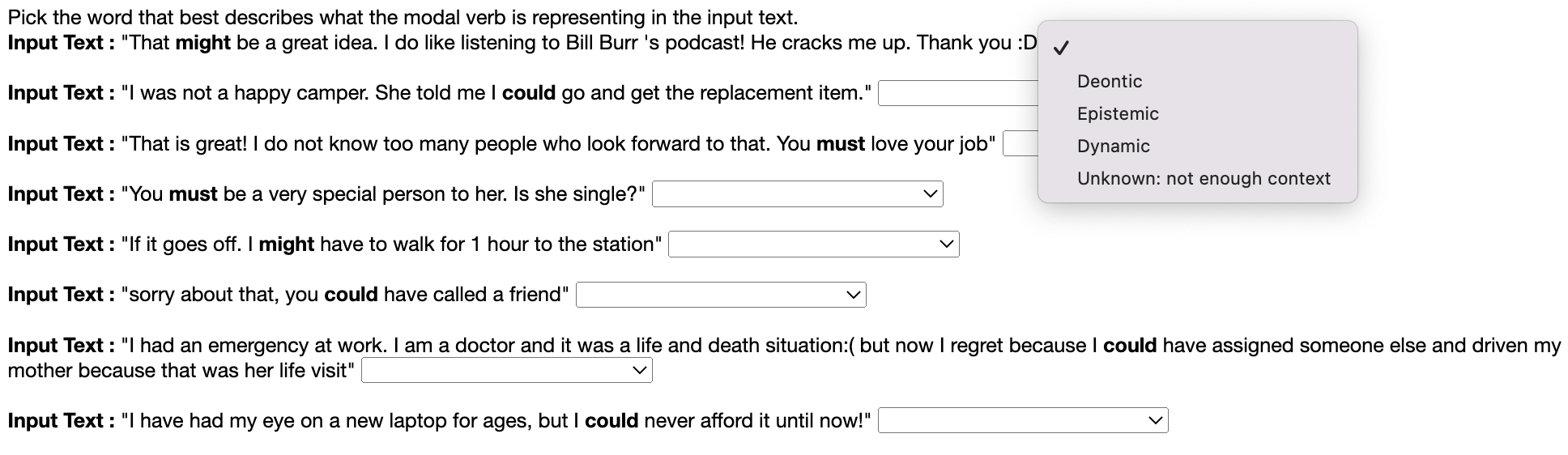}
    \captionof{figure}{Example sentences to annotate and the corresponding drop-down boxes for Palmer's categories}  
    \label{fig:mturk_sentences_cat2}
\end{minipage}

\clearpage

\begin{minipage}{1\textwidth}
    \subsection{Filtering Criteria}
    \label{sec:filtering_criteria}
    Workers were only prevented from working on further HITs when we noticed issues in their annotation quality.
    The issues were detected based on their frequency of disagreement with others and deviation from Quirk's mappings, where he laid out which labels could be assigned to which modal verbs (Table \ref{tab:quirk_categories}). We set the threshold high enough to only filter out the top 1\% of whose responses consistently deviated from both their fellow annotators and Quirk's mappings so as to not bias our data. Extreme deviation from both peers and a well-established framework implies more randomness than genuine subjective differences.\\

    \begin{center}
    \small
    \begin{tabular}{cccccc}
    \toprule
    & \smtextbfsc{\makecell[t]{can/\\could}} & \smtextbfsc{\makecell[t]{may/\\might}} & \smtextbfsc{\makecell[t]{must}} & \smtextbfsc{\makecell[t]{should}} & \smtextbfsc{\makecell[t]{will/\\would}}\\
    \midrule
    \smtextbfsc{possibility} & o & o & x & x & x\\
    \smtextbfsc{ability} & o & x & x & x & x\\
    \smtextbfsc{permission} & o & o & x & x & x\\
    \smtextbfsc{necessity} & x& x & o & x & x\\
    \smtextbfsc{obligation}& x & x & o & o & x\\
    \smtextbfsc{inference} & x & x & x & o & x\\
    \smtextbfsc{prediction} & x & x & x & x & o\\
    \smtextbfsc{volition} & x & x & x & x & o\\
    \bottomrule
    \end{tabular}\\
    \captionof{table}{Label to modal verb mapping as defined by Quirk}
    \label{tab:quirk_categories}
\vspace{0.5cm}
    \begin{tabular}{lll}
    \multicolumn{3}{r}{\# of Annotators}\\
    \toprule
    \textbfsc{\# Annotations} & \textbfsc{Quirk} & \textbfsc{Palmer}\\
    \midrule
    $< 200$ & 87 & 83\\
    200 \textasciitilde\texttt{}400 & 6 & 3\\
    400 \textasciitilde\texttt{ }600 & 3 & 1\\
    600 \textasciitilde\texttt{ }800 & 0 & 2\\
    800 \textasciitilde\texttt{ }1000 & 1 & 0\\
    1000 \textasciitilde\texttt{ }1200 & 1 & 0\\
    1200 \textasciitilde\texttt{ }1400 & 0 & 1\\
    1400 \textasciitilde\texttt{ }1600 & 0 & 1\\
    1600 \textasciitilde\texttt{ }1800 & 0 & 2\\
    1800 \textasciitilde\texttt{ }2000 & 1 & 0\\
    2000 \textasciitilde\texttt{ }2200 & 0 & 0\\
    2200 \textasciitilde\texttt{ }2400 & 1 & 0\\
    2400 \textasciitilde\texttt{ }2600 & 1 & 1\\
    \bottomrule
    \end{tabular}
    \captionof{table}{Distribution of how many annotations were contributed by each annotator}
\label{tab:annotator_contribution_distribution}
\end{center}
\end{minipage}

\clearpage
\section{Dataset Statistics}
\label{sec:dataset_statistics}
\begin{minipage}{1\textwidth}
\begin{center}
    \small
    \begin{tabular}{llll}
    \toprule
    \multicolumn{4}{c}{\textbfsc{\categoryone}}\\
    \smtextbfsc{Modal Verb} & \smtextbfsc{Agreement} & \smtextbfsc{Disagreement} & \smtextbfsc{Total}\\
    \midrule
    will  &  564  &  166  &  730 \\
    would  &  424  &  280  &  704 \\
    should  &  406  &  172  &  578 \\
    may  &  153  &  26  &  179 \\
    might  &  347  &  48  &  395 \\
    must  &  431  &  112  &  543 \\
    could  &  445  &  63  &  508 \\
    can  &  780  &  121  &  901 \\
    \midrule
    total & 3550 & 988 & 4538\\
    \midrule
    \multicolumn{4}{c}{}\\
    \multicolumn{4}{c}{\textbfsc{\categorytwo}}\\
    \smtextbfsc{Modal Verb} & \smtextbfsc{Agreement} & \smtextbfsc{Disagreement} & \smtextbfsc{Total}\\
    \midrule
    will  &  651  &  79  &  730 \\
    would  &  592  &  113  &  705 \\
    should  &  544  &  34  &  578 \\
    may  &  161  &  18  &  179 \\
    might  &  358  &  37  &  395 \\
    must  &  520  &  23  &  543 \\
    could  &  460  &  48  &  508 \\
    can  &  805  &  96  &  901 \\
    \midrule
    total & 4091 & 448 & 4539\\
    \bottomrule
    \end{tabular}
    \captionof{table}{Proportion of agreements and disagreements within the dataset. The totals do not add up to 4540 because of ``unknown'' labels, which we omitted from the table due to low count, but are included in the dataset.}
    \label{tab:agreement_disagreement_table_count}

\vspace{1cm}
    \small
    \begin{tabular}{llV{7cm}} 
    \toprule
    \textbfsc{Combination} & \textbfsc{Proportion} & \textbfsc{Example Utterances}\\
    \midrule
    inference-possibility-prediction & 8.00\%. & That \textit{should} be fun. Pokemon is a great franchise. I have many of the handheld games.\\
    \midrule
    inference-necessity-prediction & 5.16\% & The odds \textit{must} be astronomical, almost like winning the lottery.\\
    \midrule
    possibility-prediction-volition & 4.45\% & Do you mean LeBron James? I was hoping he \textit{would} come to Miami!\\
    \midrule
    ability-inference-possibility & 3.54\% & Oh no. Were you able to get things sorted out? We live far away from family and I know how hard it \textit{can} be especially when there are health concerns.\\
    \midrule
    ability-possibility-prediction & 3.44\% & True, just do not like how the world is inching toward a conflict that \textit{could} spill over to a nuclear war.\\
    \midrule
    deontic-dynamic-epistemic & 92.63\% & I was disappointed by my manager when he told that I \textit{will} probably get my promotion next year(not this year)\\
    \bottomrule
    \end{tabular}
    \captionof{table}{Top conflicting annotation triplets from \dataset}
    \label{tab:conflicting_annots}

\end{center}
\end{minipage}

\clearpage
\begin{minipage}{1\textwidth}
\begin{center}
    \includegraphics[scale=0.35]{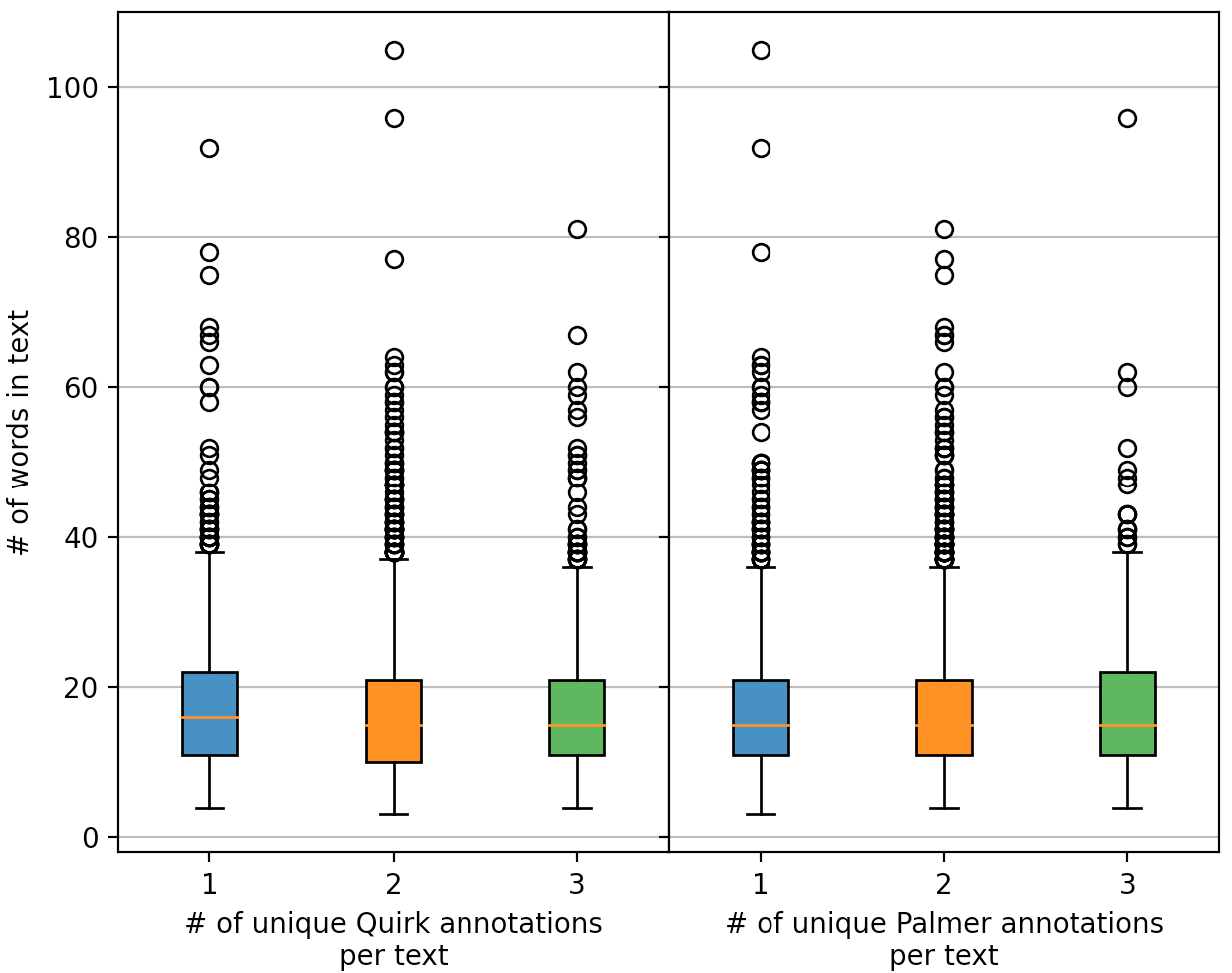}
    \captionof{table}{We see no difference in the number of unique annotations (fewer unique annotations meaning less disagreement) and the corresponding utterance length. While this is intuitively surprising, it aligns with findings from \cite{pavlick-kwiatkowski-2019-inherent}.}
    \label{fig:context_effect}
\end{center}
\vspace{1cm}
\section{Classification results}
\label{sec:classification_results}

\end{minipage}

\clearpage
\begin{minipage}{1\textwidth}
\centering
\begin{tabular}{lllll}
\toprule
\textbfsc{Model} & \textbfsc{Learning rate} & \textbfsc{Dataset} & \textbfsc{Validation F1} & \textbfsc{Test F1}\\
\midrule
ALBERT\textsubscript{base}& 5e-6 & Quirk & 75.49 & 79.36\\
BERT\textsubscript{base}& 5e-6 & Quirk & 75.02 & 77.66\\
BERT\textsubscript{large}& 5e-6 & Quirk & 77.88 & 80.56\\
RoBERTa\textsubscript{base}& 5e-6 & Quirk & 79.21 & 80.81\\
\rowcolor{Gray}
RoBERTa\textsubscript{large}& 5e-6 & Quirk & 78.98 & 82.22\\
DistilBERT\textsubscript{base}& 5e-6 & Quirk & 78.1 & 79.19\\
\midrule
ALBERT\textsubscript{base}& 1e-5 & Quirk & 69.61 & 72.67\\
BERT\textsubscript{base}& 1e-5 & Quirk & 77.84 & 78.39\\
BERT\textsubscript{large}& 1e-5& Quirk & 77.99 & 80.23\\
RoBERTa\textsubscript{base}& 1e-5& Quirk & 78.72 & 80.53\\
\rowcolor{Gray}
RoBERTa\textsubscript{large}& 1e-5 & Quirk & 78.63 & 80.62\\
DistilBERT\textsubscript{base}& 1e-5& Quirk & 77.5 & 78\\
\midrule
ALBERT\textsubscript{base}& 2e-5 & Quirk & 70.22 & 73.18\\
BERT\textsubscript{base}& 2e-5 & Quirk & 77.74 & 78.47\\
BERT\textsubscript{large}& 2e-5 & Quirk & 77.80 & 79.19\\
\rowcolor{Gray}
RoBERTa\textsubscript{base}& 2e-5 & Quirk & 78.55 & 79.88\\
RoBERTa\textsubscript{large}& 2e-5 & Quirk & 77.42 & 79.14\\
DistilBERT\textsubscript{base}& 2e-5 & Quirk & 77.02 & 77.80\\
\hline
ALBERT\textsubscript{base}& 5e-6 & Palmer & 74.66 & 75.58\\
BERT\textsubscript{base}& 5e-6 & Palmer & 76.17 & 75.49\\
BERT\textsubscript{large}& 5e-6 & Palmer & 75.22 & 75.11\\
RoBERTa\textsubscript{base}& 5e-6 & Palmer & 76.9 & 77.51\\
\rowcolor{Gray}
RoBERTa\textsubscript{large}& 5e-6 & Palmer & 77.08 & 78.36\\
DistilBERT\textsubscript{base}& 5e-6 & Palmer & 76.37 & 74.5\\
\midrule
ALBERT\textsubscript{base} & 1e-5 & Palmer & 73.63 & 74.36\\
BERT\textsubscript{base} &1e-5 & Palmer & 74.35 & 74.02\\
BERT\textsubscript{large}& 1e-5 & Palmer & 74.27 & 74.68\\
RoBERTa\textsubscript{base}& 1e-5 & Palmer & 75.94 & 76.76\\
\rowcolor{Gray}
RoBERTa\textsubscript{large}& 1e-5& Palmer & 76.09 & 76.85\\
DistilBERT\textsubscript{base}& 1e-5 & Palmer & 74.72 & 73.6\\
\midrule
ALBERT\textsubscript{base}& 2e-5 & Palmer & 74.36 & 74.79\\
BERT\textsubscript{base}& 2e-5 & Palmer & 73.66 & 72.76\\
BERT\textsubscript{large}& 2e-5 & Palmer & 73.63 & 74.16\\
\rowcolor{Gray}
RoBERTa\textsubscript{base}& 2e-5 & Palmer & 75.46 & 76.57\\
RoBERTa\textsubscript{large}& 2e-5 & Palmer & 70.54 & 70.59\\
DistilBERT\textsubscript{base}& 2e-5 & Palmer & 74.09 & 72.81\\
\bottomrule
\end{tabular}
\captionof{table}{F1 scores for fine-tuned models trained using \dataset, averaged over a 10-fold cross-validation.}
\label{tab:classifier_res2}
\end{minipage}
\clearpage
\begin{minipage}{1\textwidth}
\centering
\begin{tabular}{lllll}
\toprule
\textbfsc{Model} & \textbfsc{Learning rate} & \textbfsc{Dataset} & \textbfsc{Validation F1} & \textbfsc{Test F1}\\
\midrule
ALBERT\textsubscript{base}& 5e-6 & Palmer $\rightarrow$ \ruppenhofer & 74.26 & 47.4\\
BERT\textsubscript{base}& 5e-6 & Palmer $\rightarrow$ \ruppenhofer & 75.77 & 42.88\\
BERT\textsubscript{large}& 5e-6 & Palmer $\rightarrow$ \ruppenhofer & 75.72 & 42.29\\
RoBERTa\textsubscript{base}& 5e-6 & Palmer $\rightarrow$ \ruppenhofer & 76.89 & 52.53\\
\rowcolor{Gray}
RoBERTa\textsubscript{large}& 5e-6 & Palmer $\rightarrow$ \ruppenhofer & 76.61 & 54.78\\
DistilBERT\textsubscript{base}& 5e-6 & Palmer $\rightarrow$ \ruppenhofer & 75.74 & 47.71\\
\midrule
ALBERT\textsubscript{base}& 1e-5 & Palmer $\rightarrow$ \ruppenhofer & 71.16 & 42.09\\
BERT\textsubscript{base}& 1e-5 & Palmer $\rightarrow$ \ruppenhofer & 74.8 & 48.44\\
BERT\textsubscript{large}& 1e-5 & Palmer $\rightarrow$ \ruppenhofer & 74.57 & 50.72\\
\rowcolor{Gray}
RoBERTa\textsubscript{base}& 1e-5 & Palmer $\rightarrow$ \ruppenhofer & 75.41 & 57.99\\
RoBERTa\textsubscript{large}& 1e-5 & Palmer $\rightarrow$ \ruppenhofer & 70.47 & 57.75\\
DistilBERT\textsubscript{base}& 1e-5 & Palmer $\rightarrow$ \ruppenhofer & 74.19 & 54.58\\
\rowcolor{Gray}
\midrule
ALBERT\textsubscript{base} & 2e-5 & Palmer $\rightarrow$ \ruppenhofer & 73.64 & 52.75\\
BERT\textsubscript{base}& 2e-5 & Palmer $\rightarrow$ \ruppenhofer & 74.18 & 55.72\\
BERT\textsubscript{large}& 2e-5 & Palmer $\rightarrow$ \ruppenhofer & 74.29 & 57.4\\
\rowcolor{Gray}
RoBERTa\textsubscript{base}& 2e-5 & Palmer $\rightarrow$ \ruppenhofer & 75.4 & 61.44\\
RoBERTa\textsubscript{large}& 2e-5 & Palmer $\rightarrow$ \ruppenhofer & 70.3 & 59.1\\
DistilBERT\textsubscript{base}& 2e-5 & Palmer $\rightarrow$ \ruppenhofer & 73.7 & 57.56\\
\midrule
ALBERT\textsubscript{base} & 5e-6 & \ruppenhofer $\rightarrow$ Palmer & 83.41 & 37.08\\
BERT\textsubscript{base} & 5e-6 & \ruppenhofer $\rightarrow$ Palmer & 80.91 & 56.11\\
BERT\textsubscript{large} & 5e-6 & \ruppenhofer $\rightarrow$ Palmer & 81.35 & 52.35\\
RoBERTa\textsubscript{base}& 5e-6 & \ruppenhofer $\rightarrow$ Palmer & 85.76 & 57.15\\
\rowcolor{Gray}
RoBERTa\textsubscript{large}& 5e-6 & \ruppenhofer $\rightarrow$ Palmer & 86.5 & 66.37\\
DistilBERT\textsubscript{base} & 5e-6 & \ruppenhofer $\rightarrow$ Palmer & 82.71 & 56.36\\
\hline
ALBERT\textsubscript{base}& 1e-5 & \ruppenhofer $\rightarrow$ Palmer & 81.47 & 46.08\\
BERT\textsubscript{base}& 1e-5 & \ruppenhofer $\rightarrow$ Palmer & 81.82 & 57.23\\
BERT\textsubscript{large}& 1e-5 & \ruppenhofer $\rightarrow$ Palmer & 82.44 & 53.89\\
RoBERTa\textsubscript{base}& 1e-5 & \ruppenhofer $\rightarrow$ Palmer & 85.22 & 58.2\\
\rowcolor{Gray}
RoBERTa\textsubscript{large}& 1e-5 & \ruppenhofer $\rightarrow$ Palmer & 88.07 & 65.4\\
DistilBERT\textsubscript{base}& 1e-5 & \ruppenhofer $\rightarrow$ Palmer & 81.88 & 55\\
\midrule
ALBERT\textsubscript{base}& 2e-5 & \ruppenhofer $\rightarrow$ Palmer & 80.94 & 43.96\\
BERT\textsubscript{base}& 2e-5 & \ruppenhofer $\rightarrow$ Palmer & 82.74 & 57.13\\
BERT\textsubscript{large}& 2e-5 & \ruppenhofer $\rightarrow$ Palmer & 84.13 & 58.89\\
\rowcolor{Gray}
RoBERTa\textsubscript{base}& 2e-5 & \ruppenhofer $\rightarrow$ Palmer & 84.04 & 60.71\\
RoBERTa\textsubscript{large}& 2e-5 & \ruppenhofer $\rightarrow$ Palmer & 79.61 & 59.12\\
DistilBERT\textsubscript{base}& 2e-5 & \ruppenhofer $\rightarrow$ Palmer & 80.45 & 57.36\\
\rowcolor{Gray}
\bottomrule
\end{tabular}
\captionof{table}{Observing cross-domain transferability between \categorytwo and Ruppenhofer and Rehbein (\ruppenhofer). We see a clear performance domination of the RoBERTa models.}
\label{tab:classifier_res4}



\end{minipage}